\documentclass[letterpaper, 10 pt, conference]{ieeeconf}  

\IEEEoverridecommandlockouts                              

\usepackage{graphicx} 
\usepackage{subcaption}
\usepackage{hyperref}
\usepackage[nolist]{acronym}
\usepackage{siunitx}
\usepackage{gensymb}

\usepackage{cite}

\usepackage[skip=8pt,font=small]{caption}

\renewcommand{\baselinestretch}{0.99}

\title{\LARGE \bf
Collecting Human Motion Data in Large and Occlusion-Prone Environments using Ultra-Wideband Localization
}

\author{Janik Kaden$^{1}$, Maximilian Hilger$^{2}$, Tim Schreiter$^{2,3}$, Marius Schaab$^{2}$, \\Thomas Graichen$^{4}$, Andrey Rudenko$^{5}$, Ulrich Heinkel$^{1}$, and Achim J. Lilienthal$^{2,3}$
\thanks{$^{1}$Chemnitz University of Technology, Chair for Circuit and System Design, Chemnitz, Germany,
        {\tt\small {firstname.name}@etit.tu-chemnitz.de}}%
\thanks{$^{2}$Munich Institute of Robotics and Machine Intelligence (MIRMI), Technical University of Munich, Germany,
        {\tt\small {firstname.name}@tum.de}}%
\thanks{$^{3}$Center for Applied Autonomous Sensor Systems (AASS), Örebro, Sweden,
        {\tt\small {firstname.name}@oru.se}}%
\thanks{$^{4}$Pinpoint GmbH, Chemnitz, Germany,
        {\tt\small thomas.graichen@pinpoint.de}}%
\thanks{$^{5}$Bosch Corporate Research, Germany,
        {\tt\small andrey.rudenko@de.bosch.com}}%
\thanks{This work was supported by the EU Horizon 2020 No. 101017274 (DARKO) and by the Federal Ministry of Education and Research (BMBF) within the project ELFE as part of the WIR! program.}%
}

\begin{document}

\acrodef{slam}[SLAM]{Simultaneous Localization and Mapping}
\acrodef{uwb}[UWB]{Ultra-Wideband}
\acrodef{imu}[IMU]{Inertial Measurement Unit}

\maketitle
\thispagestyle{empty}
\pagestyle{empty}

\begin{abstract} 
With robots increasingly integrating into human environments, understanding and predicting human motion is essential for safe and efficient interactions. 
Modern human motion and activity prediction approaches require high quality and quantity of data for training and evaluation, usually collected from motion capture systems, onboard or stationary sensors. 
Setting up these systems is challenging due to the intricate setup of hardware components, extensive calibration procedures, occlusions, and substantial costs. 
These constraints make deploying such systems in new and large environments difficult and limit their usability for in-the-wild measurements.
In this paper we investigate the possibility to apply the novel \ac{uwb} localization technology as a scalable alternative for human motion capture in crowded and occlusion-prone environments. 
We include additional sensing modalities such as eye-tracking, onboard robot LiDAR and radar sensors, and record motion capture data as ground truth for evaluation and comparison.
The environment imitates a museum setup, with up to four active participants navigating toward random goals in a natural way, and offers more than 130 minutes of multi-modal data. 
Our investigation provides a step toward scalable and accurate motion data collection beyond vision-based systems, laying a foundation for evaluating sensing modalities like UWB in larger and complex environments like warehouses, airports, or convention centers. 

\end{abstract}

\section{INTRODUCTION}
\label{sec:introduction}
Understanding human motion is a cornerstone for intelligent robots to interact seamlessly with humans in shared spaces. 
Recent advances in human motion prediction do not only use geometric and velocity information but also leverage semantic and contextual cues for more accurate performance~\cite{Rudenko.2020}. 
These approaches often depend on substantial amounts of high-quality data for training and evaluation, e.g., generated using a motion capture system \cite{schreiterTHORMAGNILargescaleIndoor2024} or high-resolution LiDAR sensors \cite{Ehsanpour.2022}. 
Acquiring this data is costly and limited by the volume covered by the motion capture system or by occlusions in the LiDAR data. 
To generate larger-scale datasets, it is necessary to establish new ways of collecting human motion data. 
Ultra-Wideband (\ac{uwb}) technology is an increasingly adopted solution for precise indoor localization in the consumer market, as a growing number of smartphone manufacturers integrate UWB hardware into their devices \cite{qorvoQorvoDeliversUltraWideband2021,estimoteWhyApples2nd2023}. 
In 2023, native UWB positioning on a smartphone was demonstrated for the first time \cite{qorvoQorvoDemonstratesUWB2023}. 

With the untracked navigation use case standardized by the FiRa consortium\footnote{\url{https://www.firaconsortium.org/}} in their 2.0 specification, GPS-comparable solutions for smartphones with indoor accuracy in the decimeter range are now possible. This standardization enables UWB device interoperability and supports a broad adoption and integration across consumer and industrial applications. The underlying Downlink Time-Difference-of-Arrival method allows scalable and private positioning directly on the user's device. Accuracy can be improved by fusing with additional device sensors, such as an \ac{imu}. 
Due to standardization and growing demand for precise indoor navigation, the availability of UWB-enabled smartphones will continue to increase. This will provide new digital experiences in venues like museums, which until now have relied on specially tailored solutions based on less accurate technologies like Bluetooth \cite{kyudertsedenovIndoorNavigationComprehensive2024}.

\begin{figure}[!t]
\centering
\begin{subfigure}[b]{0.17\textwidth}
	\includegraphics[width = 1\textwidth]{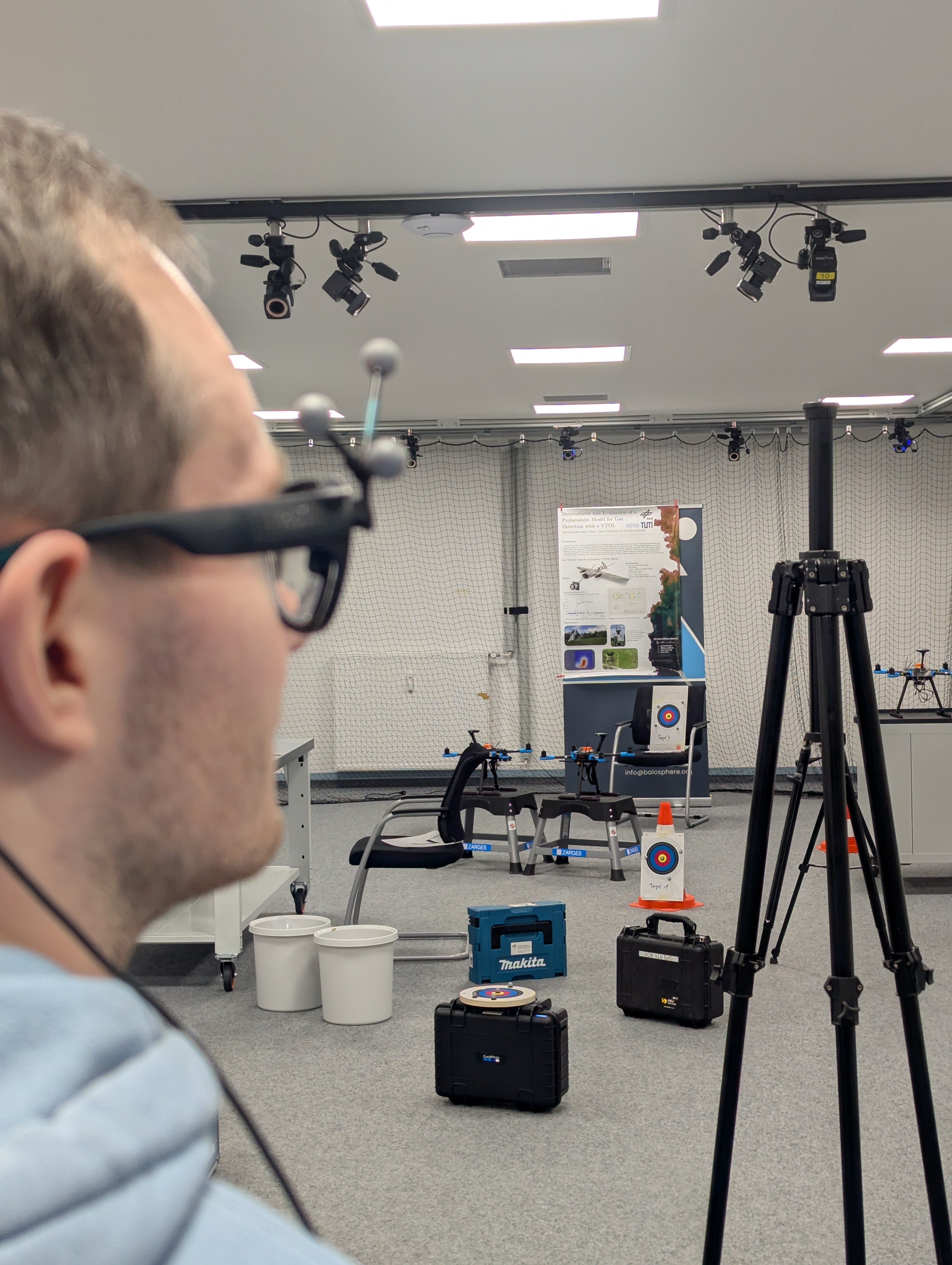}
	\caption{Environment}
   \end{subfigure}
\begin{subfigure}[b]{0.2971\textwidth}
    \includegraphics[width=1\textwidth]{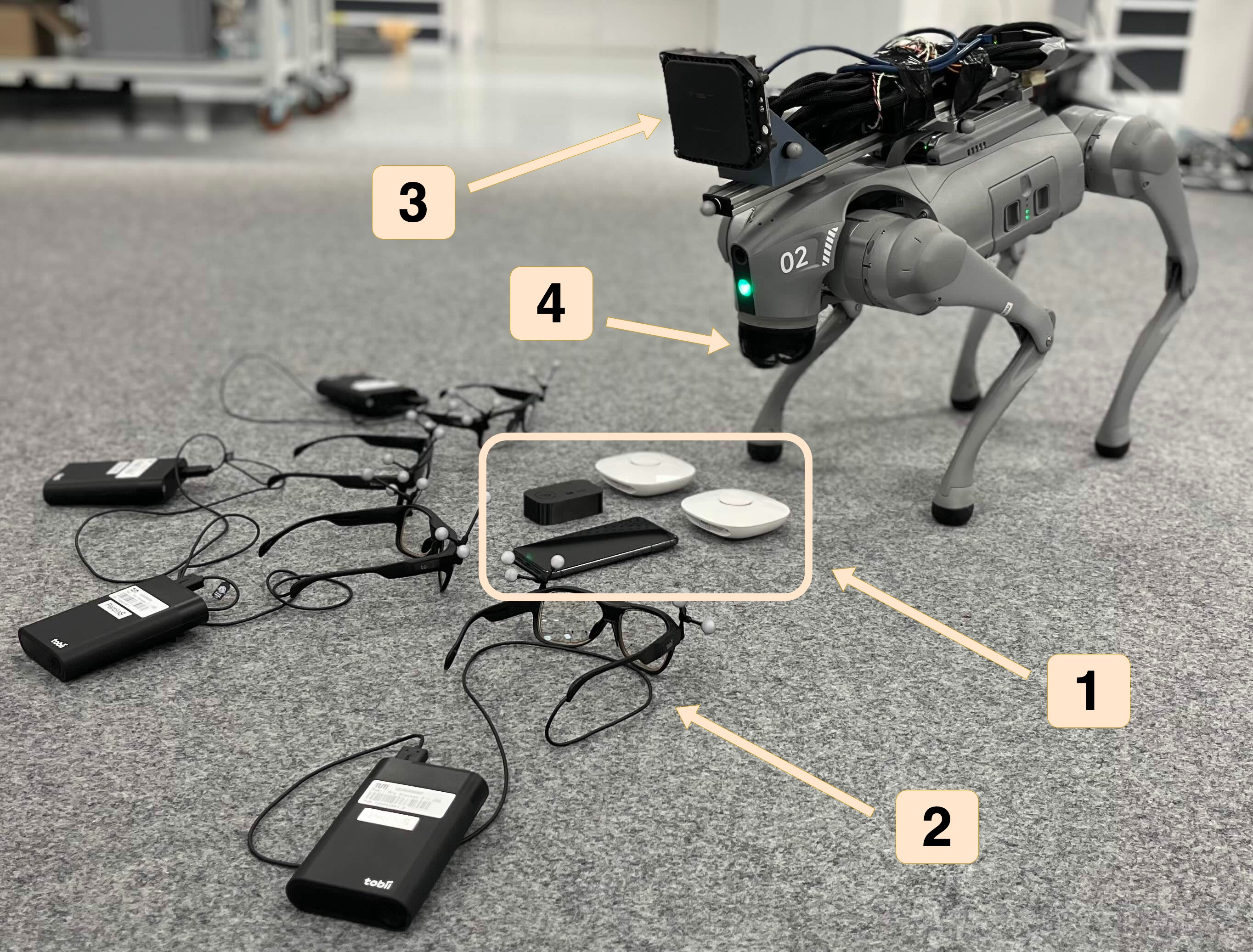}
    \caption{Modalities}
\end{subfigure}
\caption{\textbf{Overview of the datasets modalities and recording environment:} The UWB system (1), eye-tracking glasses (2) and the robot with radar (3) and LiDAR (4) sensors.}
\label{modalities}
\vspace{-3mm}
\end{figure}

In this work, we investigate extension of the THÖR protocol for human motion data collection \cite{rudenkoTHORHumanRobotNavigation2020} to include UWB tracking of moving people. 
THÖR features a scripted indoor environment to generate goal-driven and natural human motion in crowded social spaces containing fixed obstacles and a moving robot. 
The participants draw random cards at the goal points, indicating the next target point. 
THÖR-MAGNI \cite{schreiterTHORMAGNILargescaleIndoor2024} further extends this protocol by enriching the environment with semantic contexts, such as areas of caution or one-way passages, and adds diverse tasks and activities for people, including several modes of interaction with the robot. It contains over 3.5 hours of motion data for 40 participants. 
Both THÖR and THÖR-MAGNI are open source\footnote{\url{http://thor.oru.se}} and are used to improve human motion prediction algorithms \cite{Rudenko.2020}, e.g., based on deep learning \cite{dealmeidaTHORMagniComparativeAnalysis2023}, causal discovery \cite{castri2022causal}
or physics-based methods \cite{rudenko2022atlas}. In both datasets, 3D LiDAR data is also available.

In this paper, we extend the THÖR data collection setup from an industrial to a public museum environment. 
Museum layouts are designed to attract the visitor's visual attention \cite{krukarWalkLookRemember2014,harveyInfluenceMuseumExhibit1998}. 
Hence, the obstacle setup in our dataset intends to steer the participants' visual attention similarly, which we can quantify by recording eye-tracking data. 
The dataset features multiple goal points (museum exhibits) and diverse static obstacles in the room, encouraging natural human motion behavior. 
In addition, a robot equipped with both LiDAR and radar sensors is utilized. 
The availability of velocity measurements and the penetration abilities makes radar an interesting sensing modality in dynamic and occluded environments.
Uniquely for this recording, 2D UWB trajectories are available for up to three participants at the same time. Finally, accurate ground truth positions are recorded by a motion capture system.

The rest of the paper is organized as follows: Sec.~\ref{section:data_collection} describes the recording modalities and the interaction scenarios. 
We show preliminary results in Sec.~\ref{section:recorded_data}. 
After completing post-processing and data curation, the data of the \ac{uwb}-localization, the robot's sensors, the motion capture system, and the raw eye-tracking data will be made available\footnote{\url{https://doi.org/10.5281/zenodo.15211243}}. 

\section{DATA COLLECTION}
\label{section:data_collection}

\subsection{Room Setup} 
\label{subsec:room_setup}

The data collection occurred in a robot lab at the German Aerospace Center (DLR). 
The main test area measures \qty{15.64}{\meter} $\times$ \qty{6.68}{\meter} and is covered by the motion capture system and \ac{uwb}. 
Six goal positions are marked within the test area. 
In some experiments, this room is complemented by a second room not covered by the motion capture system. 
This introduces an area without optical tracking to demonstrate the use of UWB as a standalone localization method. 
The room simulates a temporary exhibition space, similar to those commonly seen in museums where installations are frequently changed. 
We implement this setup in two layouts.

\begin{figure}[!t]
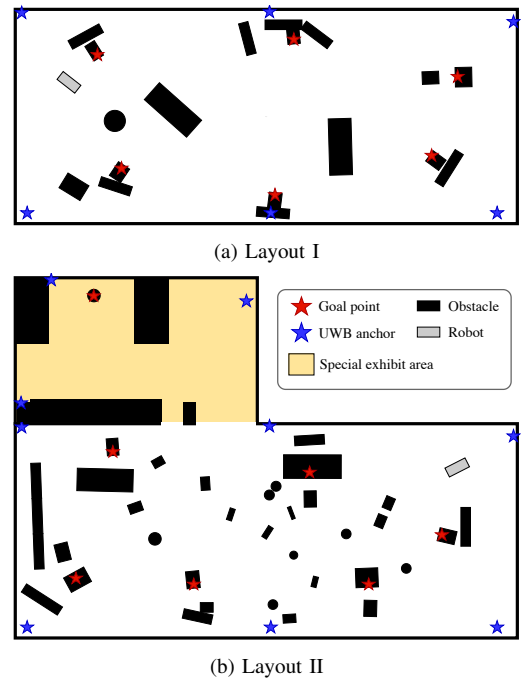

\centering
\begin{subfigure}[b]{0.38\textwidth}
	\includegraphics[width = 1\textwidth]{images/obstacle_layout_1_uwb_robot.drawio.pdf}
	\caption{Layout I}
    \label{subfig:layout1_map}
    \vspace{1 mm} 
\end{subfigure}
\begin{subfigure}[b]{0.38\textwidth}
	\includegraphics[width = 1\textwidth]{images/obstacle_layout_2_uwb_robot_legend.drawio.pdf}
	\caption{Layout II}
    \label{subfig:layout2_map}
\end{subfigure}
\caption{\textbf{Top-down view of the indoor museum room setup.} Layout II also shows the additional ``Special Exhibit'' area.} 
\label{fig:hdmaps}
\vspace{-3mm}
\end{figure}

The obstacles in Layout I are designed to encourage the visual search behavior of participants' gaze. 
Two large tables block parts of the room; some of the goals are situated behind roll-ups and do not provide a direct line of sight of parts of the room. 
Fig.~\ref{subfig:layout1} shows a picture of the obstacle setup, and Fig.~\ref{subfig:layout1_map} depicts a map of the room layout. 
In Layout II, in contrast, many small obstacles are placed in the middle of the room to encourage navigating motion patterns. 
This forces the participants to choose between multiple different paths. 
A picture of the second layout is given in Fig.~\ref{subfig:layout2}, and Fig.~\ref{subfig:layout2_map} illustrates a map of the layout. 
The robot is placed in a corner of the room in all the layouts mentioned above, except for the ``Moving Robot'' Scenario 6 introduced in Sec.~\ref{subsec:scenarios}. 

\subsection{Scenario Description}
\label{subsec:scenarios}

In total, seven scenarios in two obstacle layouts are recorded to capture various movements and interactions. 
Each scenario includes multiple runs with varying numbers of participants, and each run is 3–5 minutes. 
The participants are assigned a random goal point with a deck of cards at the start of each run. 
Each participant draws one card indicating the next random goal target, returns it to the bottom of the card deck, and proceeds to the next goal point.
The card decks are designed to favor longer and more complex paths between goal points.  

Starting with an empty room, in Scenario 1, the participants are instructed to build up Layouts I or II, which were given by a floor map drawing. 
The participants thus not only have the Visitor role used in \cite{schreiterTHORMAGNILargescaleIndoor2024} and \cite{rudenkoTHORHumanRobotNavigation2020} but also have a similar role to the Carrier role where boxes had to be transported. 
After building up the layouts, regular test runs are conducted in Scenarios 3 and 4 with one to four simultaneous participants. 
Scenario 2 depicts the participants' regular (baseline) motion, where only the goal points are present in the volume.
The ``Special Exhibit'' Scenario 5 includes the additional room depicted in Fig.~\ref{subsec:room_setup}. 
The quadruped robot moves through the area in Scenario 6. 
After recording all runs for the layouts, the participants were instructed to deconstruct the layouts in Scenario 7.

\subsection {Sensing modalities}
\label{subsec:modalities}
The dataset features a unique combination of sensing modalities. 
Our sensor setup is selected such that two claims can be verified prior to the larger-scale recordings: i) human motion data collection is possible with \ac{uwb}, and ii) environment reconstruction in dynamic and occluded environments can be accomplished in a combination of radar and LiDAR. 
Motion capture data serves as accurate ground truth.
\subsubsection{UWB Indoor Positioning} 

\begin{figure}[!t]
\centering
\begin{subfigure}[b]{0.24035\textwidth}
    \includegraphics[width = 1\textwidth]{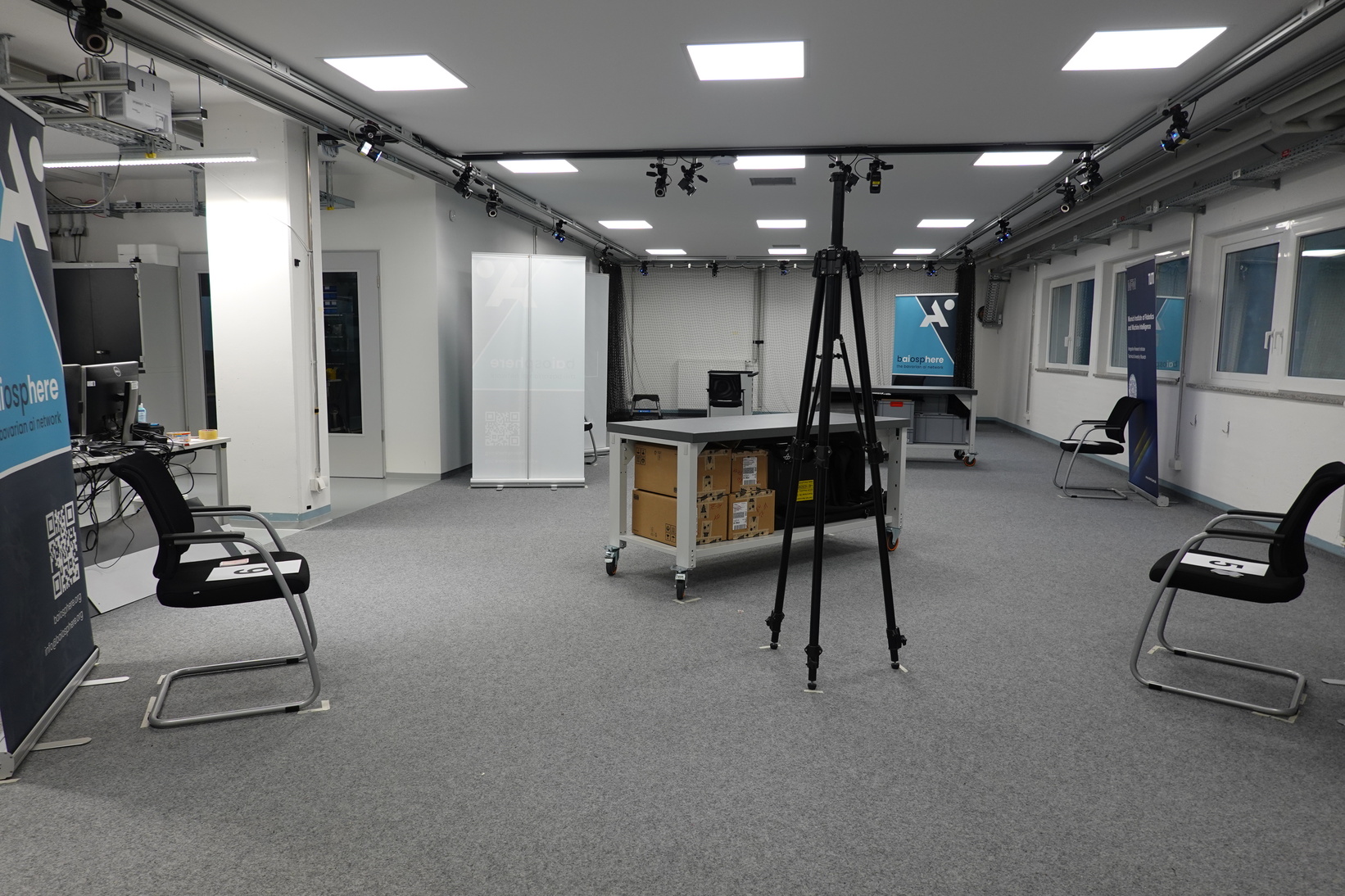}
	\caption{Layout I}
    \label{subfig:layout1}
\end{subfigure}
\begin{subfigure}[b]{0.21375\textwidth}
    \includegraphics[width=1\textwidth]{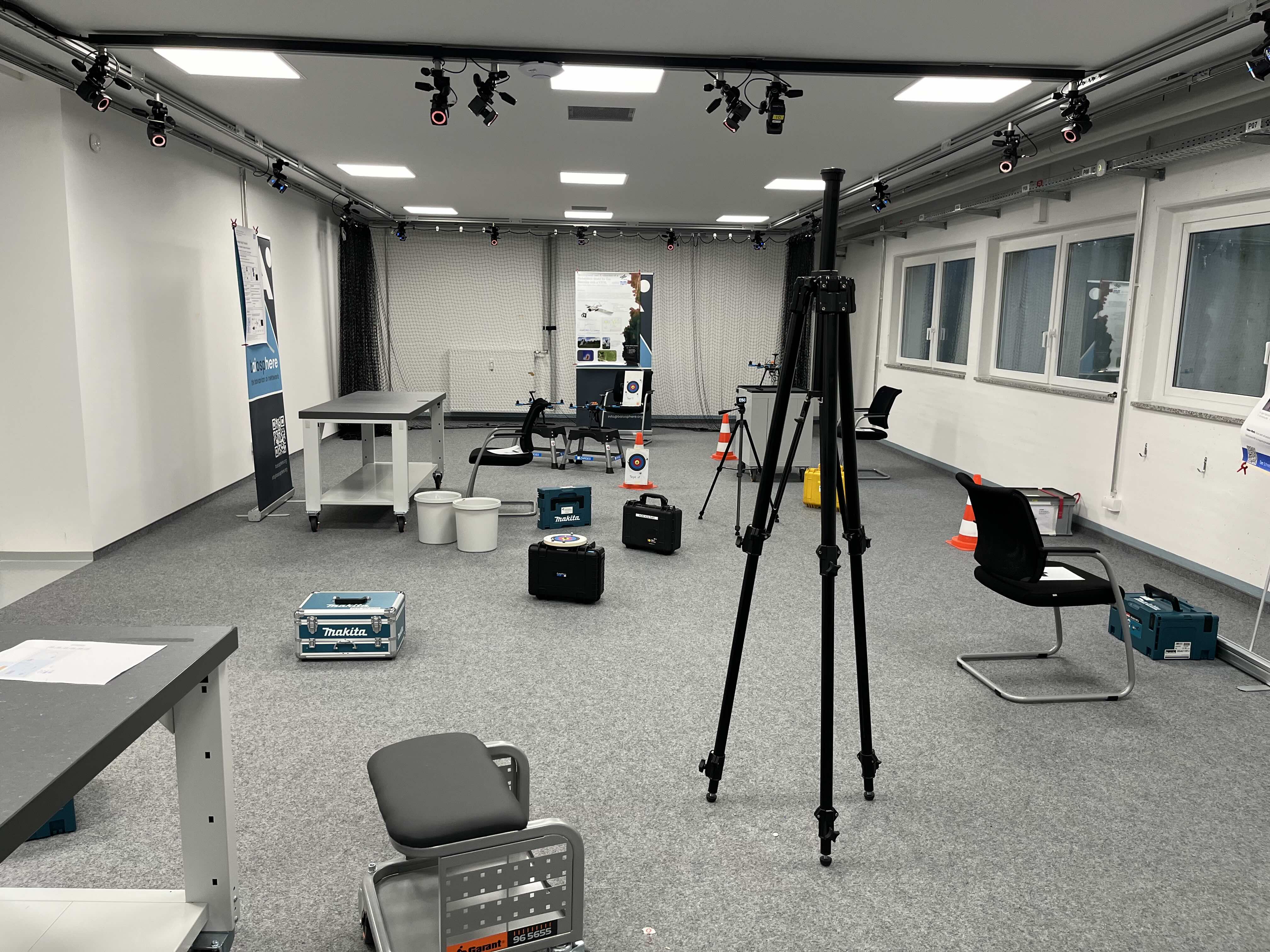}
    \caption{Layout II}
    \label{subfig:layout2}
\end{subfigure}
\caption{\textbf{The two obstacle layouts used in the recordings.} They feature a few large obstacles that need to be navigated around (a) and many small obstacles that can be stepped over (b). Retractable banners introduce occlusions in the environment, adding visual search behavior to the scenario.}
\label{fig:scenarios}
\vspace{-3mm}
\end{figure}

The UWB system is a FiRa 2.0 compliant indoor localization system by the German company Pinpoint GmbH\footnote{\url{https://pinpoint.de/en}}. 
In total, nine anchors (called SATlets) are installed at known locations in the room to provide localization coverage. 
These are comparable to GNSS satellites that send out information that the receivers use to calculate their own position. 
For the main area seen in Fig.~\ref{fig:scenarios}, six SATlets were placed on the room's walls at a height of about \qty{2.5}{\meter}. 
For dedicated ``UWB-only'' runs, which included a special exhibit outside the motion capture coverage, three SATlets were added to cover this area. 
The placement is shown in Fig.~\ref{fig:hdmaps}, depicted by the blue stars. The UWB anchors are battery-powered and synchronize wirelessly, eliminating the need to run cables throughout the room and significantly reducing setup time. After manually measuring the room with a laser distance meter having an accuracy of ± \qty{3}{\milli\meter}, the Pinpoint app ``EasyPlan'' was used to configure the SATlets. This includes the position in the local coordinate system and some UWB-specific physical layer (PHY) parameters, such as the channel or preamble. Previous work has shown that the UWB PHY impacts the performance of UWB connections between devices \cite{kadenPerformanceInvestigationIEEE2024}. Lower-frequency configurations generally support longer communication ranges, while higher-frequency settings offer better resilience against interference from other wireless technologies, such as WiFi 6E. Given the short distances in our exhibit area, we selected a FiRa-compatible configuration optimized for robustness against potential interference.

Three different UWB receiver devices were used to capture the trajectory of the participants: a TRACElet (battery-powered tag) and two smartphones (Samsung Galaxy S24+ and Google Pixel 8 Pro). The TRACElet was attached to a lanyard the participants wore during the runs, and the smartphones were handheld. The recorded 2D UWB positions represent the raw data obtained directly from the devices at \qty{4}{\hertz}, without additional sensor fusion.

\begin{figure}[!t]
\centering
    \begin{subfigure}[b]{0.165\textwidth}
		\includegraphics[width = 1\textwidth]{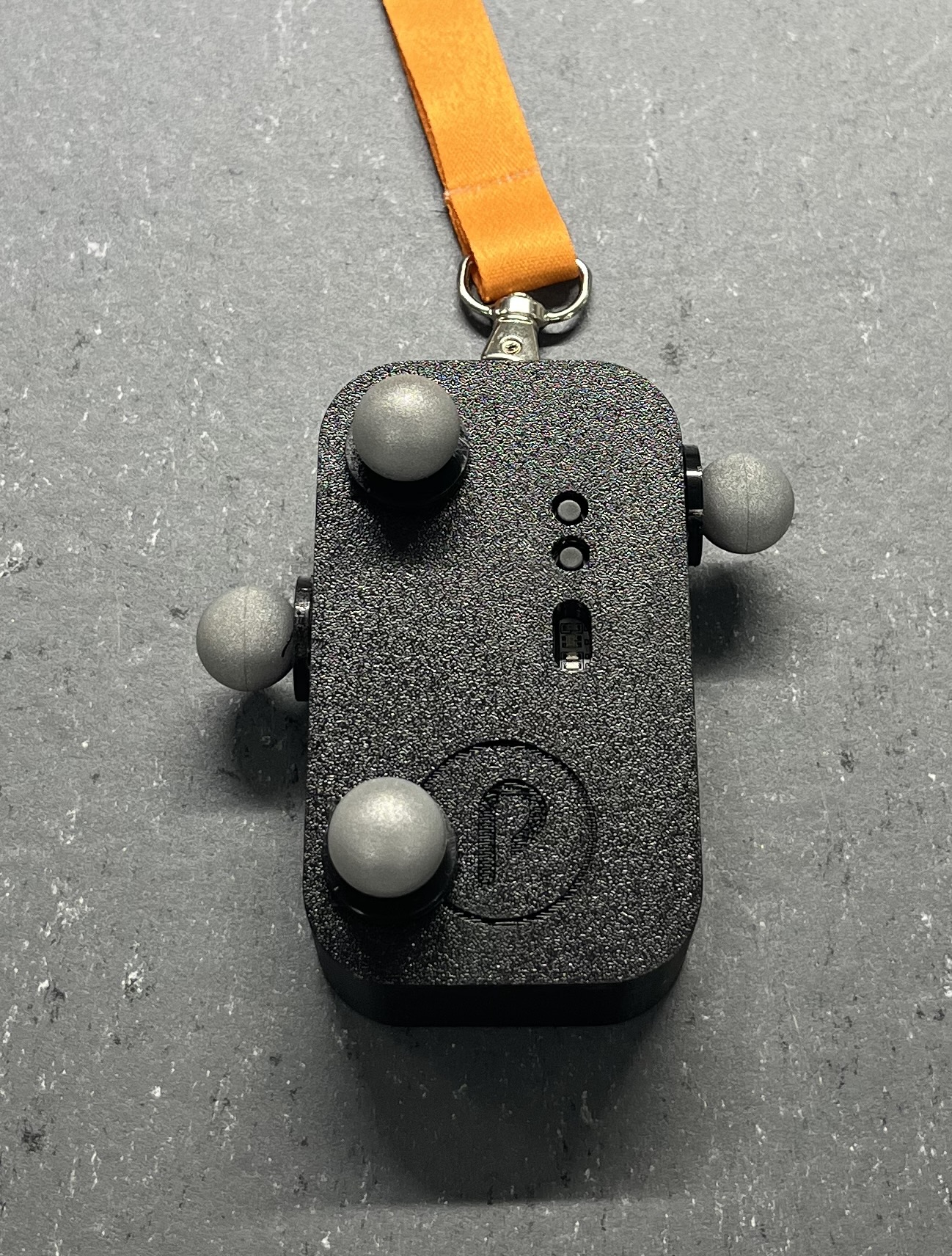}
		\caption{TRACElet}
    \end{subfigure}
    \begin{subfigure}[b]{0.18\textwidth}
		\includegraphics[width = 1\textwidth]{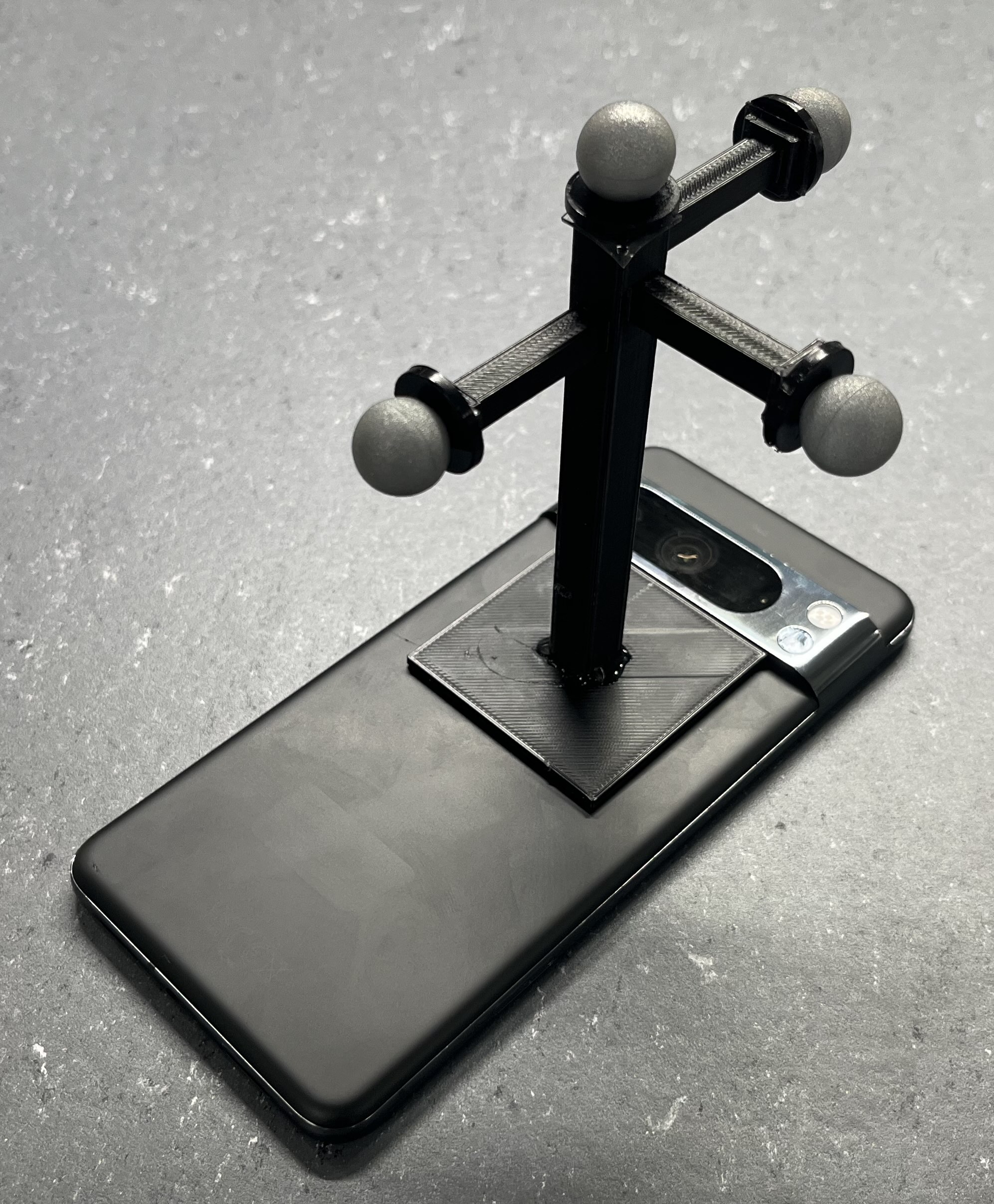}
		\caption{Google Pixel 8 Pro}
    \end{subfigure}
\caption{\textbf{Two UWB receiver devices used in the runs.} Devices are equipped with IR-Markers for 6DoF MotionCapture}
\label{figure:uwb_devices}
\end{figure}

\subsubsection{Eye tracking}
\label{subsubsec:eye_tracking}

Four wearable eye-tracking glasses (Tobii Pro Glasses 3) were used to capture the gaze information of the participants. 
Each pair of glasses consists of a head unit worn like a regular pair of glasses. 
If the participant needed a visual aid, corrective lenses were attached. 
We used the Tobii Motion Capture Marker sets and the Vicon integration to simultaneously track the participants and record the gaze information using the Vicon Tracker software\footnote{\url{https://help.vicon.com/space/Tracker310}}. 
This ensures the data is synchronized and the gaze information can be used in the motion capture coordinate system. The glasses with the recording units and the motion capture marker sets attached are shown in Fig.~\ref{modalities}.

\subsubsection{Robot with LiDAR and radar}
The dataset features a UniTree Go2 quadruped robot,  equipped with a UniTree L1 LiDAR with built-in \ac{imu} and a Bosch Off-Highway Premium radar. 
The low-cost LiDAR has a non-repetitive scan pattern. 
It produces 21.6k points per second, substantially lower than the 2.6m points per second available in similar datasets \cite{schreiterTHORMAGNILargescaleIndoor2024}.
The sparsity of the data makes human motion tracking and \ac{slam} in dynamic settings more challenging because less geometric context is available to detect humans.
This challenge can be addressed using our proposed data collection setup. 
The radar sensor outputs a sparse point cloud comprising 3D geometrical information and radial velocity obtained through the Doppler effect. 
In particular, the velocity information can be leveraged for human motion tracking, for mapping of dynamics \cite{Kucner.2017}, and to detect moving objects in \ac{slam}. 
Additionally, radar waves can penetrate through some materials, which benefits mapping in occluded environments.
The maps estimated by \ac{slam}, in turn, can be used to understand the recorded human trajectory data, especially if semantic attributes are deducted from the point clouds during \cite{Chen.2019} or after mapping \cite{Sun.2018}.
The point clouds and \ac{imu} measurements were recorded in ROS2 bag files. 

\subsubsection{Motion capture}

A Vicon motion capture system is used to obtain a highly accurate ground truth for the participants, the robot, and the UWB devices. 
It consists of 26 cameras installed in the ceiling of the room. 
These cameras were calibrated prior to the recordings and covered the entire volume of the recording room, including the participants' eye-tracking glasses and the robot on the floor. 
All tracked objects are equipped with IR markers, as seen in Fig.~\ref{modalities} and Fig.~\ref{figure:uwb_devices}, for 6 DoF-based tracking of rigid bodies with the motion capture software. 
This software supports the direct integration of the Tobii eye-tracking glasses. 
Hence, the motion capture system recordings include the poses of the rigid bodies and the position of the left and right eye, pupil diameter, left and right gaze, and gaze position for all participants. 
The data is stored in a CSV format.

\subsection{Recording Procedure}
\label{subsec:recording_procedure}

One experimenter operated the motion capture, UWB, and robot recording software to ensure the recording quality and check the status of all systems. 
We followed a precise workflow for each recording to have reproducible results. 
An eye tracker calibration routine was carried out for each participant by looking into a calibration card to achieve the best possible results. 
A successful calibration was indicated in the motion capture system. The recording of the motion capture, UWB, and robot systems was then started.

\begingroup
\renewcommand{\arraystretch}{1.1}

\begin{table}[]
\centering
\caption{Overview of all scenarios}
\label{table:scenarios}
\begin{tabular}{|c|l|l|l|}
\hline
\textbf{Scenario} & \textbf{Description} & \textbf{Participants} & \textbf{Time recorded} \\
\hline
1 & Build-up & $3$ & $\qty{16}{\min} + \qty{6}{\min}$ \\
2 & Baseline & $1 - 4$ &  $8 \times \qty{3}{\min}$ \\ 
3 & Layout I & $1 - 4$ & $8 \times \qty{3}{\min}$ \\ 
4 & Layout II & $1 - 4$ & $8 \times \qty{3}{\min}$ \\ 
5 & Special Exhibit & $3$ & $6 \times \qty{4}{\min}$ \\
6 & Moving Robot & $3$ & $2 \times \qty{5}{\min}$ \\
7 & Deconstruction & $3$ & $2 \times \qty{4}{\min}$ \\
\hline
\textbf{Total} & & & \qty{136}{\min} \\
\hline
\end{tabular}
\end{table}
\endgroup

\section{RECORDED DATA}
\label{section:recorded_data}

\begin{figure}[!t]
\centering
\begin{subfigure}[b]{0.225\textwidth}
    \includegraphics[trim=0.5cm 6.2cm 0.5cm 6.2cm, width = 1\textwidth]{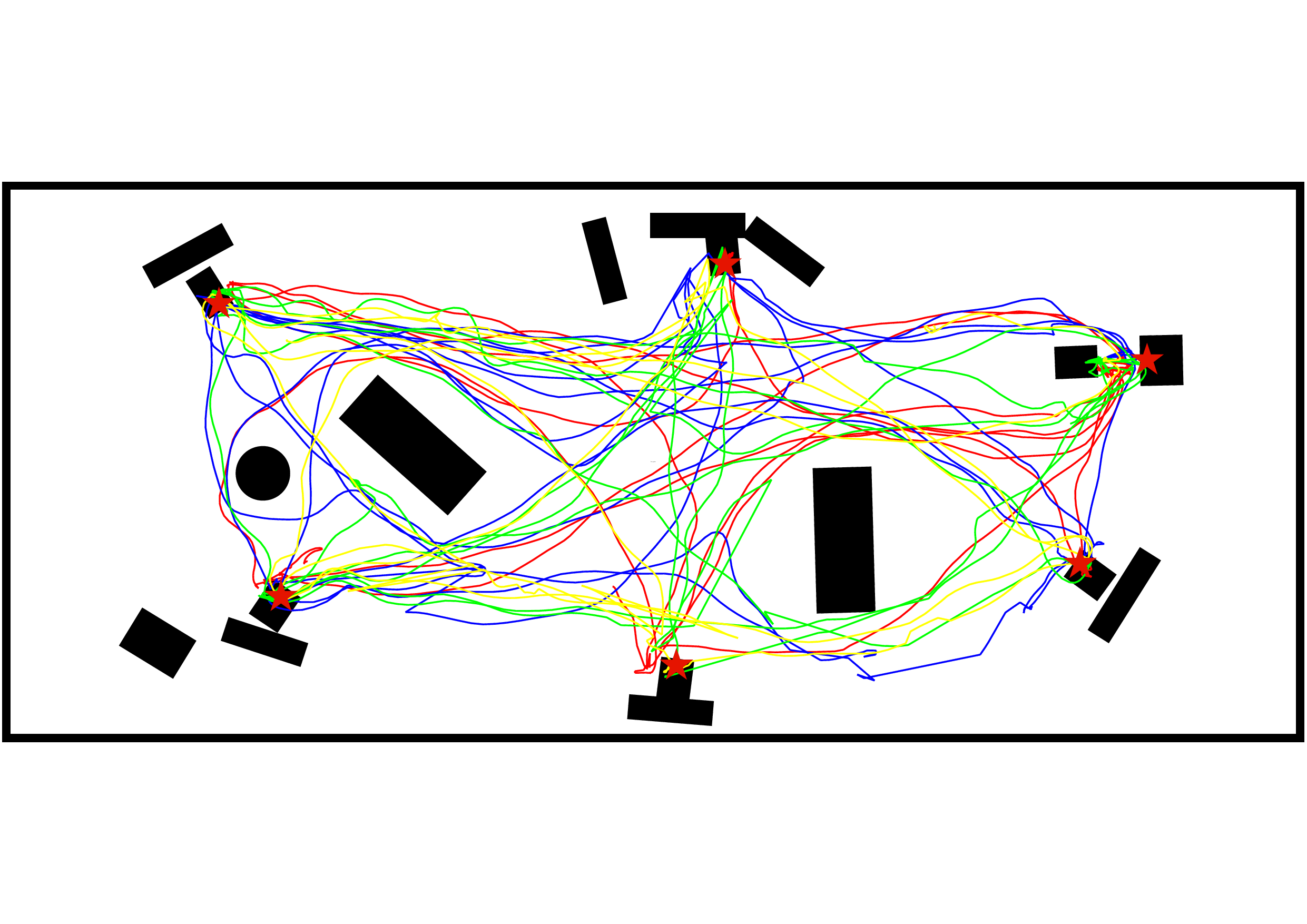}
    \caption{Scenario 3 (Layout I)}
    \label{subfig:layout1_traj}
\end{subfigure}
\begin{subfigure}[b]{0.225\textwidth}
	\includegraphics[trim=0.5cm 6.2cm 0.5cm 6.2cm, width = 1\textwidth]{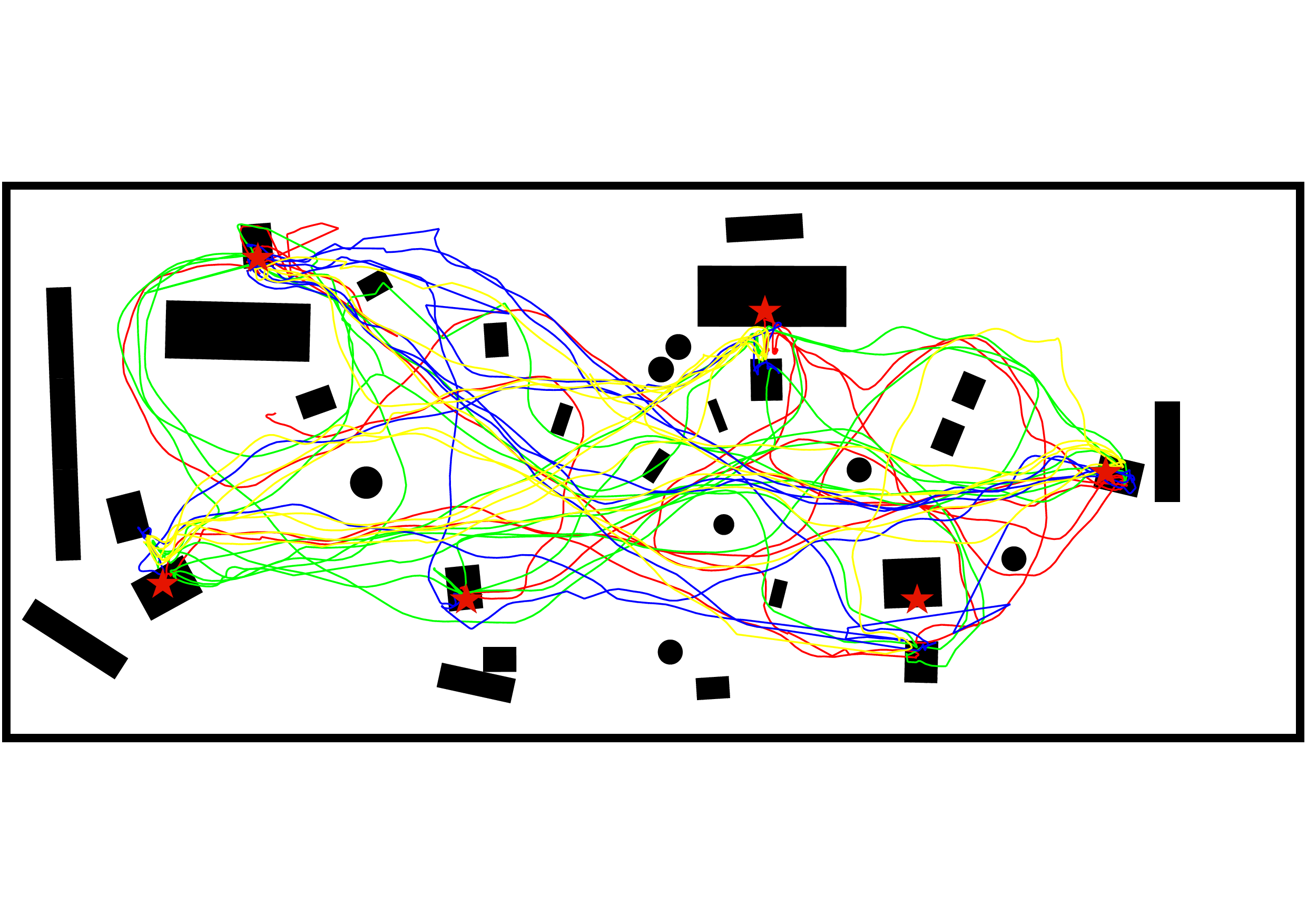}
	\caption{Scenario 4 (Layout II)}
    \label{subfig:layout2_traj}
\end{subfigure}
\caption{\textbf{Trajectories of four participants, recorded with the motion capture system.} Results of a 3-minute run, colors corresponding to unique participants.}
\label{fig:trajectories} 
\end{figure}

As described in Sec.~\ref{subsec:scenarios}, each recording lasts at least 3 minutes. 
The values for each scenario are given in Table~\ref{table:scenarios}. 
This results in over 130 minutes of multi-modal data: eye-gaze data from up to four eye-tracking glasses, 2D UWB trajectories, poses of the motion capture system, and radar and LiDAR point clouds captured by the robot.
Fig.~\ref{fig:trajectories} shows the trajectories in both layouts with four participants and a static robot. 
It is visible that the higher number of obstacles leads to more complex trajectories between the goals.

\begin{figure}[!t]
\centering
\includegraphics[width=0.4\textwidth]{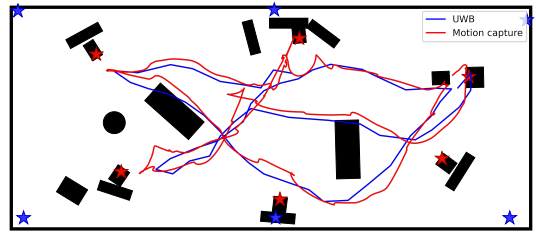}
\caption{\textbf{Top-down view of the trajectories recorded by the motion capture and the UWB system.} The figure illustrates 1 minute of movement from one participant holding the Google Pixel smartphone in hand during Scenario 3.}
\label{fig:uwb_trajectory}
\end{figure}

To evaluate the accuracy of the UWB-based localization, we compare the trajectories recorded by the motion capture system and the UWB system. 
We calculate the mean 2D displacement error using the root mean square error to quantify the system's precision. 
Fig.~\ref{fig:uwb_trajectory} shows 1 minute of movement in Scenario 3, recorded using the Pixel smartphone and having an average error of \qty{41}{\centi\meter}. 

\begin{figure}[!t]
\centering
\begin{subfigure}[b]{0.285\textwidth}
    \includegraphics[width = 1\textwidth]{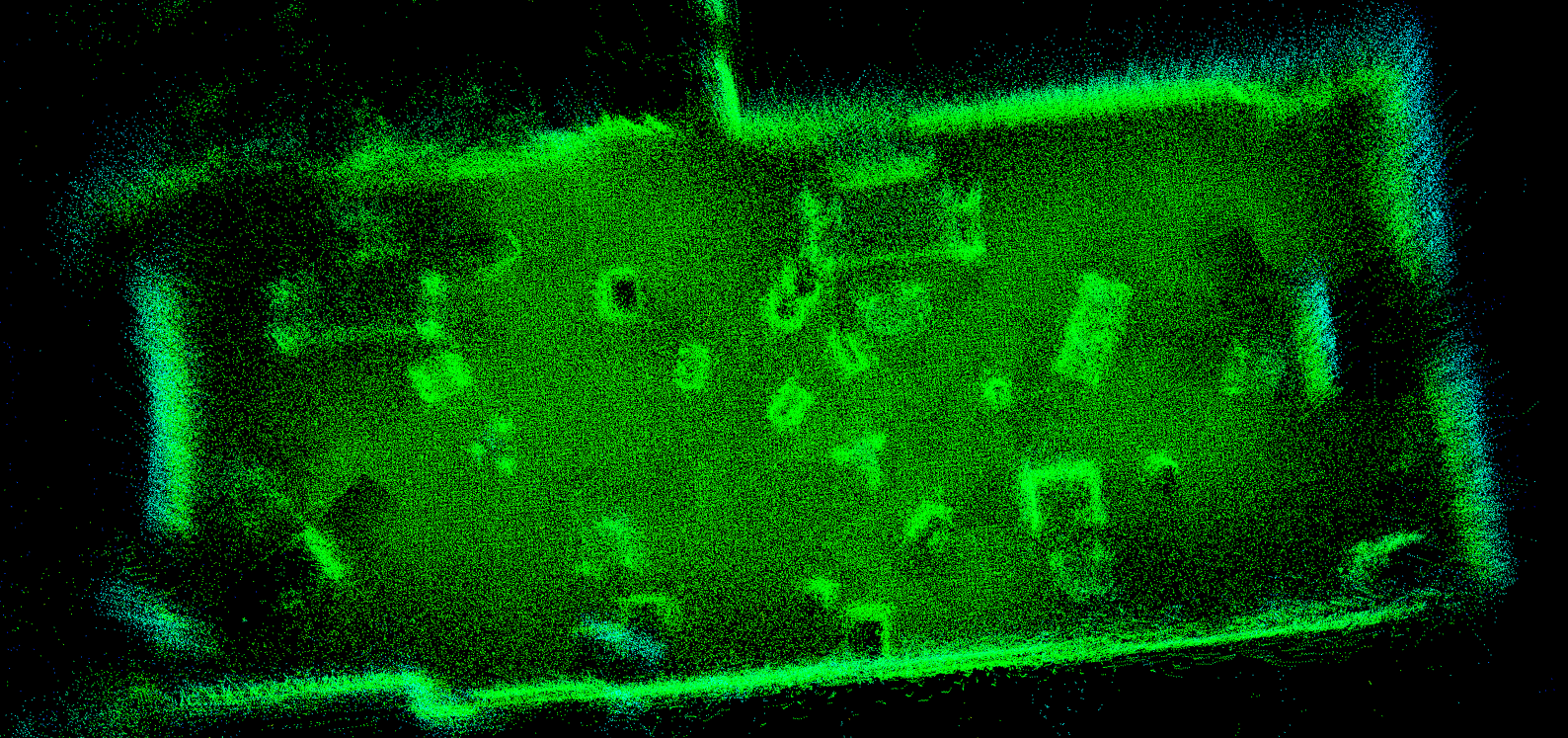}
    \caption{LiDAR map}
    \label{subfig:map_LiDAR}
\end{subfigure}
\begin{subfigure}[b]{0.19\textwidth}
	\includegraphics[width = 1\textwidth]{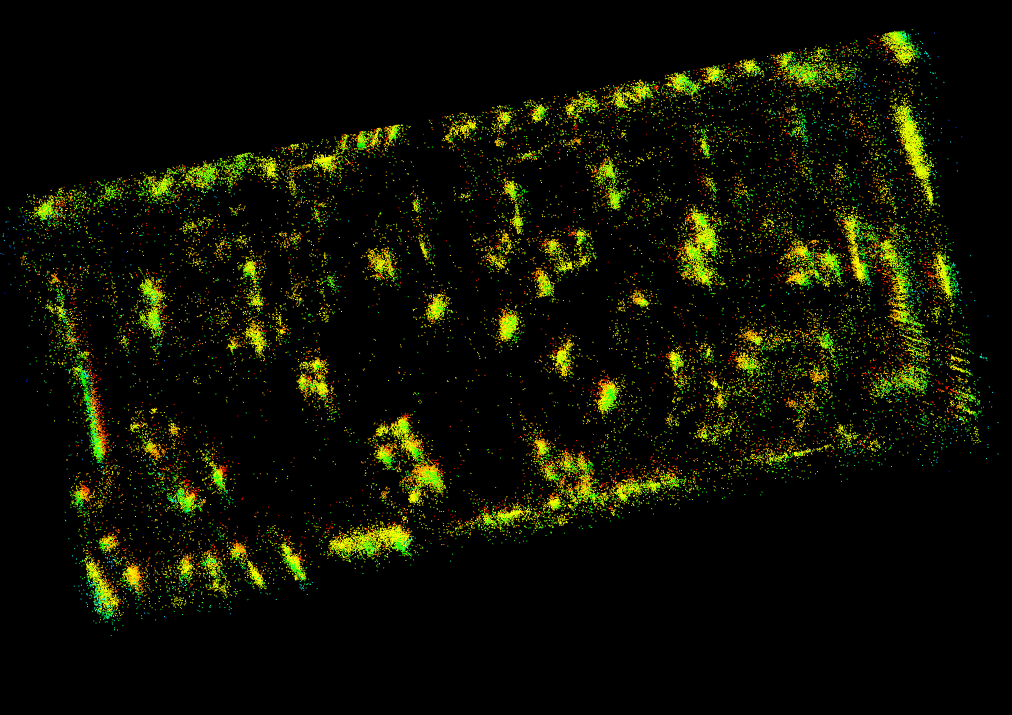}
	\caption{Radar map}
    \label{subfig:map_radar}
\end{subfigure}
\caption{\textbf{Point cloud maps of Layout II generated based on ground truth trajectories.} The LiDAR map captures more geometric details compared to the radar map.}
\label{fig:maps}
\vspace{-3mm}
\end{figure}

To showcase the mapping capabilities of the robots' sensors, we used the trajectories recorded with the motion capture system in Scenario 4 to accumulate LiDAR and radar points in a voxel grid. 
We filter out dynamic points belonging to moving participants using the tracked eye-tracking glasses. 
For the radar point cloud, noise points outside the experimental volume are discarded.
Fig.~\ref{fig:maps} depicts the recorded maps. 
Even with the sparse measurements, the LiDAR map shows accurate geometry. 
Compared to that, the radar map is more sparse and contains no ground reflections. 
Future work will investigate if the same maps can be estimated solely from the robot's onboard sensors.

\section{CONCLUSION AND FUTURE WORK}
\label{section:conclusion}

We present a novel set of human motion recordings in a simulated contextually rich indoor museum-like environment. The dataset features a unique set of technological modalities combining motion capture, UWB indoor localization, eye-tracking glasses, radar, LiDAR, and a moving quadruped robot. 
With that, we pave the way toward large-scale dataset recordings in real-world settings, leveraging consumer hardware. 
Future work will thoroughly evaluate the performance of the UWB localization system compared to the ground truth motion capture. Furthermore, leveraging the potential of UWB people tracking and 3D environment reconstruction from on-board robot sensors, we aim to investigate the possibility to collect the gaze data in crowded environments without motion capture systems.

\addtolength{\textheight}{-0cm}   





\section*{ACKNOWLEDGMENT}
The authors would like to thank the German Aerospace Center (DLR) for supporting this work by providing their facilities. 
In particular, we thank Thomas Wiedemann and Patrick Hinsen for their support during the dataset recording.
We also thank Martin Magnusson for the fruitful discussions.


\bibliographystyle{IEEEtran}
\bibliography{IEEEabrv,bibliography/bibliography}

\end{document}